\documentclass[runningheads]{llncs}
\usepackage{graphicx}
\usepackage[disable]{todonotes}
\usepackage{amsmath}
\usepackage{amssymb}
\usepackage{tikz}
\usepackage{pgfplots}
\usepackage{url}

\pgfplotsset{compat=1.11,
        /pgfplots/ybar legend/.style={
        /pgfplots/legend image code/.code={%
        \draw[##1,/tikz/.cd,bar width=10pt,yshift=-0.4em,bar shift=0pt]
                plot coordinates {(0cm,1em)};},
},
}

\usepackage{listings}
\usepackage{color}

\definecolor{commentcolor}{rgb}{0.6,0,0}
\definecolor{stringcolor}{rgb}{0.95,0.4,0}
\definecolor{keywordcolor}{rgb}{0,0.4,0}

\lstnewenvironment{pythonlisting}{\vspace{1em}}{\vspace{1em}}

\begin{document}
\title{Software and application patterns for explanation methods}
%
%
\author{Maximilian Alber}
\authorrunning{M. Alber}
%
\institute{\email{maximilian.alber@tu-berlin.de}\\ TU Berlin, Germany}
\maketitle              
\begin{abstract}
  Deep neural networks successfully pervaded many applications domains and are increasingly used in critical decision processes.
  Understanding their workings is desirable or even required to further foster their potential as well as to access sensitive domains like medical applications or autonomous driving.
  One key to this broader usage of explaining frameworks is the accessibility and understanding of respective software.
  In this work we introduce software and application patterns for explanation techniques that aim to explain individual predictions of neural networks.
  We discuss how to code well-known algorithms efficiently within deep learning software frameworks
  and describe how to embed algorithms in downstream implementations.
  Building on this we show how explanation methods can be used in applications
  to understand predictions for miss-classified samples, to compare algorithms or networks, and to examine the focus of networks.
  Furthermore, we review available open-source packages and discuss challenges posed by complex and evolving neural network structures to explanation algorithm development and implementations.
\keywords{Machine Learning  \and Artificial Intelligence \and Explanation \and Interpretability \and Software}
\end{abstract}
\section{Introduction}
\label{secintroduction}

Recent developments showed that neural networks can be applied successfully in many technical applications like computer vision~\cite{krizhevsky2012imagenet,he2016deep,lecun2015deep}, speech synthesis~\cite{van2016wavenet} and translation~\cite{vaswani2017attention,sutskever2014sequence,bahdanau2014neural}.
Inspired by such successes many more domains use machine learning and specifically deep neural networks for, e.g., material science and quantum physics~\cite{montavon2013machine,schutt2017schnet,schutt2017quantum,chmiela2017machine,chmiela2018towards}, cancer research~\cite{binder2018towards,korbar2017looking}, strategic games~\cite{silver2016mastering,silver2017mastering}, knowledge embeddings~\cite{mikolov2013distributed,pennington2014glove,alber2017kernel}, and even for automatic machine learning~\cite{zoph2017learning,alber2018backprop}.
With this broader application focus the requirements beyond predictive power alone rise.
One key requirement in this context is the ability to understand and interpret predictions made by a neural network or generally by a learning machine.
In at least two areas this ability plays an important role:
domains that require an understanding because they are intrinsically critical or because it is mandatory by law,
and domains that strive to extract knowledge beyond the predictions of learned models.
As exemplary domains can be named: health care~\cite{binder2018towards,korbar2017looking,gondal2017weakly}, applications affected by the GDPR~\cite{voigt2017eu}, and natural sciences~\cite{montavon2013machine,schutt2017schnet,schutt2017quantum,chmiela2017machine}.

The advancement of deep neural networks is due to their potential to leverage complex and structured data by learning complicated inference processes.
This makes a better understanding of such models challenging, yet a rewarding target.
Various approaches to tackle this problem have been developed, e.g.,~~\cite{ancona2018towards,poener18,montavon2018methods,lundberg2017unified,selvaraju2017grad}.
While the nature and objectives of explanation algorithms can be ambiguous~\cite{lipton2016mythos},
in practice gaining specific insights can already enable practitioners and researchers to create knowledge as first promising results show,
e.g.,~\cite{LapCVPR16,LapAMFG17,sebastianNatComm,binder2018towards,zintgraf2017visualizing,sundararajan2017axiomatic}.

To facilitate the transition of explanation methods from research
into wide- spread application domains
the existence and understanding of standard usage patterns and software is of particular importance.
This, on one hand, lowers the application barrier and effort for non-experts and,
on the other hand, it allows experts to focus on algorithm customization and research.
With this in mind, this chapter is dedicated to the software and application patterns for implementing and using explanation methods with deep neural networks.
In particular we focus on the explanation techniques that have in common to highlight features in the input space of a targeted neural network~\cite{MonPR17}.

In the next section we address this by a step-by-step showcasing on how explanation methods can be realized efficiently and highlight important design patterns.
The final part of the section shows how to tune the algorithms and how to visualize obtained results.
In Section~\ref{secapplications} we extend this by integrating explanation methods in several generic application cases with the aim to understand predictions for miss-classified samples, to compare algorithms or networks, and to examine the focus of networks.
The remainder, Section~\ref{secsoftwarepackages},~\ref{secchallenges} and~\ref{secconclusion}, addresses available open-source packages, further challenges and gives a conclusion.



\section{Implementing explanation algorithms}
\label{secimplementation}

\begin{figure}[t]
  \centering
  \includegraphics[width=0.95\textwidth]{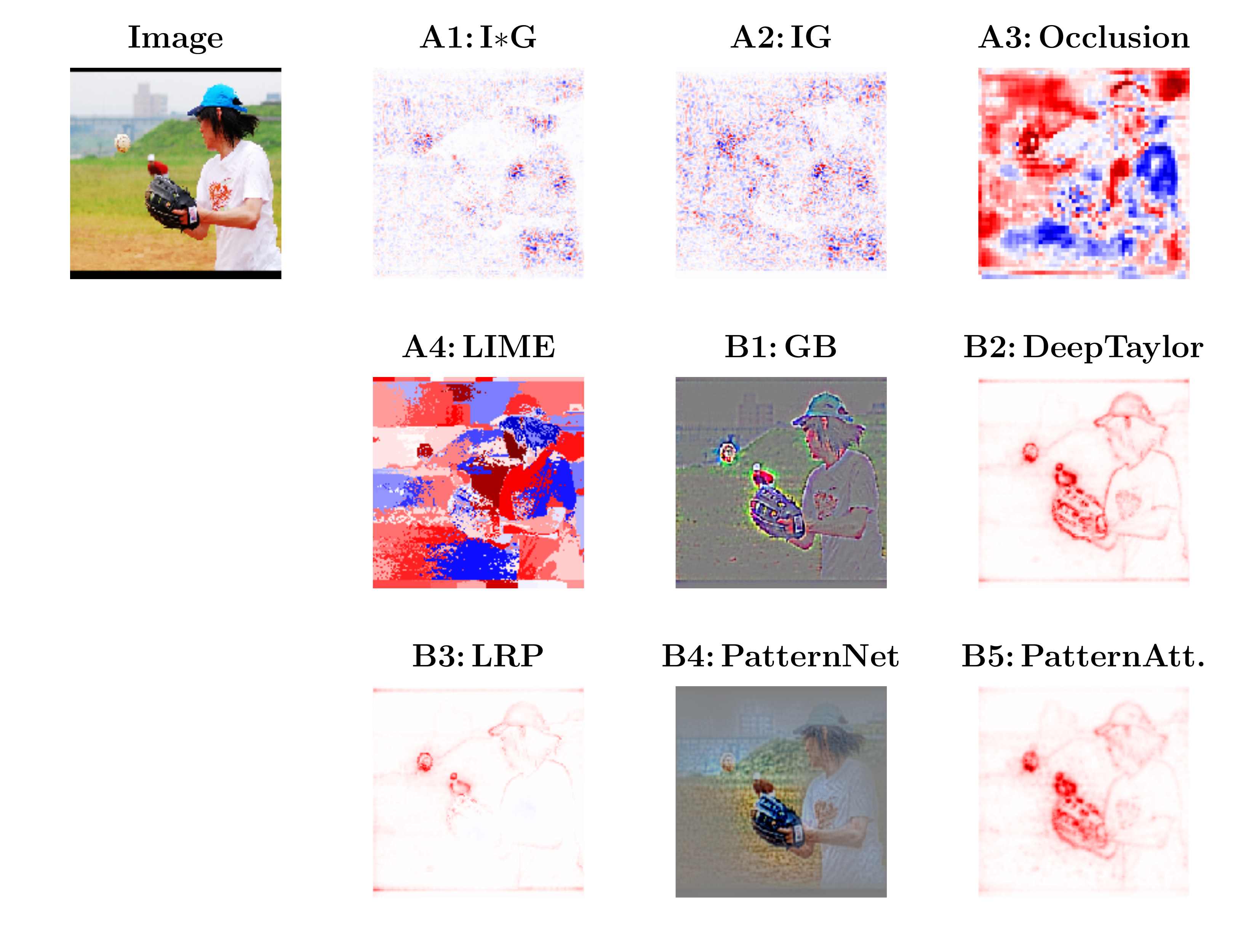}
  \caption{\textbf{Exemplary application of the implemented algorithms:} This figure shows the results of the implemented explanation methods applied on the image in the upper-left corner using a VGG16 network~\cite{simonyan2014very}.
    The prediction- or gradient-based methods (group A) are Input * Gradient~\cite[A1]{kindermans2016investigating,shrikumar2017learning}, Integrated Gradients~\cite[A2]{sundararajan2017axiomatic}, Occlusion~\cite[A3]{zeiler2014visualizing}, and LIME~\cite[A4]{ribeiro2016should}.
    The propagation-based methods (group B) are Guided Backprop~\cite[B1]{springenberg2015striving}, Deep Taylor~\cite[B2]{MonPR17}, LRP~\cite[B3]{LapAMFG17}, PatternNet \& PatternAttribution~\cite[B4 and B5]{kindermans2018learning}.
    On how the explanations are visualized we refer to Section~\ref{secvisualizing}.
    Best viewed in digital and color.
  }
  \label{figimplemented}
\end{figure}

Implementing a neural network efficiently can be a complicated and error-prone process and
additionally implementing an explanation algorithm makes things even trickier.
We will now introduce the key patterns of explanation algorithms that allow for an efficient and structured implementation.
Subsequently we complete the section by explaining how to approach interface design, parameter tuning, and visualization of the results.

To make the code examples as useful as possible we will not rely on pseudo-code, but rather use Keras~\cite{chollet2015keras}, TensorFlow~\cite{abadi2016tensorflow} and iNNvestigate~\cite{alber2019innvestigate} to implement our examples for the example network VGG16~\cite{simonyan2014very}.
The results are illustrated in Figure~\ref{figimplemented} and will be created step-by-step.
The code listings contain the most important code fragments and we provide corresponding executable code as Jupyter notebook at \url{https://github.com/albermax/interpretable_ai_book__sw_chapter}.

Let us recall that the algorithms we explore have a common functional form,
namely they map from the input to a equal-dimensional saliency map,
e.g., the output saliency map has the same tensor shape as the input tensor.
More formal: given a neural network model that maps some input to a single output neuron $f: \mathbb{R}^n \mapsto \mathbb{R}$,
the considered algorithms have the following form $e: \mathbb{R}^n \mapsto \mathbb{R}^n$.
We will select as output neuron the neuron with the largest activation in the final layer.
Any other neuron could also be used.
We assume that the target neural network is given as Keras model and
the corresponding input and output tensor are given as follows:

\newpage
\begin{pythonlisting}
# Create model without trailing softmax
model = make_a_keras_model()

# Get TF tensors
input, output = model.inputs[0], model.outputs[0]
# Reduce output to response of neuron with largest activation
max_output = tf.reduce_max(output, axis=1)

# Select a sample image
x_not_pp = select_a_sample_image()
# and preprocess it for the network
x = preprocess(x_not_pp)
\end{pythonlisting}

The explanation algorithms of interest can be divided into two major groups depending on how they treat the given model.
The first group of algorithms uses only the model function or gradient to extract information about the model's prediction process by repetitively calling them with altered inputs.
The second group performs a custom backpropagation along the model graph, i.e., requires the ability to introspect the model and adapt to its composition.
Methods of the latter are typically more complex to implement,
but aim to gain insights more efficiently and/or of different quality.
The next two subsections will describe implementations for each group respectively.

\subsection{Prediction- and gradient-based explanations}

Algorithms that only rely on function or on 
gradient evaluations can be
of very simple, yet effective nature~\cite{kindermans2016investigating,shrikumar2017learning,smilkov2017smoothgrad,sundararajan2017axiomatic,zeiler2014visualizing,springenberg2015striving,zintgraf2017visualizing,ribeiro2016should}.
A downside can be the their runtime, which is often a multiple of a single function call.

\paragraph{Input * gradient} As a first example we consider input * gradient~\cite{kindermans2016investigating,shrikumar2017learning}.
The name already says it: the algorithm consists of an element-wise multiplication of the input times the gradient.
The corresponding formula is:
\begin{equation}
  e(x) = x \odot \nabla_x f(x).
\end{equation}
The method can be implemented as follows and the result is marked as A1 in Figure~\ref{figimplemented}:

\begin{pythonlisting}
# Take gradient of output neuron w.r.t. to the input
gradient = tf.gradients(max_output, input)[0]
# and multiply it with the input
input_t_gradient = input * gradient
# Run the code with TF
A1 = sess.run(input_t_gradient, {input: x})
\end{pythonlisting}

\paragraph{Integrated Gradients} A more evolved example is the method Integrated Gradients~\cite{sundararajan2017axiomatic} which tries to capture the effect
of non-linearities better by computing the gradient along a line between input image and a given reference image $x'$.
The corresponding formula for $i$-th input dimension is:
\begin{equation}
  e(x_i) = (x_i-x'_i) \odot \int_{\alpha=0}^1 \frac{\delta f(x)}{\delta x_i}\Bigr|_{\substack{x=x'+\alpha (x-x')}} d\alpha.
\end{equation}
To implement the method the integral is approximated with a finite sum 
and, building on the previous code snippet, the code looks as follows (result is tagged with A2 in Figure~\ref{figimplemented}):

\begin{pythonlisting}
# Nr. of steps along path
steps = 32
# Take as reference a black image,
# i.e., lowest number of the networks input value range.
x_ref = np.ones_like(x) * net['input_range'][0]
# Take gradient of output neuron w.r.t. to input
gradient = tf.gradients(max_output, input)[0]

# Sum gradients along the path from x to x_ref
gradient_sum = np.zeros_like(x)
for step in range(steps):
  # Create intermediate input
  x_step = x_ref + (x - x_ref) * step / steps
  # Compute and add the gradient for intermediate input
  gradient_sum += sess.run(gradient, {input: x_step})

# Integrated Gradients formula
A2 = gradient_sum * (x - x_ref)
\end{pythonlisting}

\paragraph{Occlusion} In contrast to the two presented methods occlusion-based methods rely on the function value instead of its gradient, e.g.,~\cite{zeiler2014visualizing,li2016understanding,zintgraf2017visualizing}.
The basic variant~\cite{zeiler2014visualizing} divides the input, typically an image, into a grid of non-overlapping patches.
Then each patch gets the function value assigned
that is obtained when the patch region in the original image is perturbed or replaced by a reference value.
Eventually all values are normalized with the default activation given when no patch is occluded.
The algorithm can be implemented as follows and the result is denoted as A3 in Figure~\ref{figimplemented}:

\begin{pythonlisting}
diff = np.zeros_like(x)
# Choose a patch size
psize = 8

# Occlude patch by patch and calculate activation for each patch
for i in range(0, net['image_shape'][0], psize):
  for j in range(0, net['image_shape'][0], psize):

    # Create image with the patch occluded
    occluded_x = x.copy()
    occluded_x[:, i:i+psize, j:j+psize, :] = 0

    # Store activation of occluded image
    diff[:, i:i+psize, j:j+psize, :] = sess.run(
      max_output, {input: occluded_x})[0]

# Normalize with initial activation value
A3 = sess.run(max_output, {input: x})[0] - diff
\end{pythonlisting}

\paragraph{LIME} The last prediction-based explanation class, e.g.,~\cite{ribeiro2016should,lundberg2017unified}, decomposes the input sample into features.
Subsequently, prediction results for inputs --- composed of perturbed features --- are collected,
yet instead of using the values directly for the explanation, they are used to learn an importance value for the respective features.

One representative algorithm is ``Local interpretable model-agnostic explanations''~\cite[LIME]{ribeiro2016should} that learns a local regressor for each explanation.
It works as follows for images.
First the image is divided into segments, e.g., continuous color regions.
Then a dataset is sampled where the features are randomly perturbed, e.g., filled with a gray color.
The target of the sample is determined by the prediction value for the accordingly altered input.
Using this dataset a weighted, regression model is learned and the resulting weight vector's values indicate the importance of each segment in the neural network's initial prediction.
The algorithm can be implemented as follows and the result is denoted as A4 in Figure~\ref{figimplemented}:

\begin{pythonlisting}
# Segment (not pre-processed) image
segments = skimage.segmentation.quickshift(
  x_not_pp[0], kernel_size=4, max_dist=200, ratio=0.2)
nr_segments = np.max(segments)+1


# Create dataset
nr_samples = 1000
# Randomly switch segments on and off
features = np.random.randint(0, 2, size=(nr_samples, nr_segments))
features[0, :] = 1

# Get labels for features
labels = []
for sample in features:
  tmp = x.copy()
  # Switch segments on and off
  for segment_id, segment_on in enumerate(sample):
    if segment_on == 0:
      tmp[0][segments == segment_id] = (0, 0, 0)
  # Get predicted value for this sample
  labels.append(sess.run(max_output, {input: tmp})[0])


# Compute sample weights
distances = sklearn.metrics.pairwise_distances(
            features,
            features[0].reshape(1, -1),
            metric='cosine',
).ravel()
kernel_width = 0.25
sample_weights = np.sqrt(np.exp(-(distances ** 2) / kernel_width ** 2))

# Fit L1-regressor
regressor = sklearn.linear_model.Ridge(alpha=1, fit_intercept=True)
regressor.fit(features, labels, sample_weight=sample_weights)
weights = regressor.coef_


# Map weights onto segments
A4 = np.zeros_like(x)
for segment_id, w in enumerate(weights):
  A4[0][segments == segment_id] = (w, w, w)
\end{pythonlisting}

As initially mentioned a drawback of prediction- and gradient-based methods can be slow runtime, which is often a multiple of a single function evaluation --- as the loops in the code snippets already suggested.
For instance Integrated Gradients used $32$ evaluations, the occlusion algorithm $(224/4)^2 = 56^2 = 3136$ and LIME $1000$ (same as in~\cite{ribeiro2016should}).
Especially for complex networks and for applications with time constraints this can be prohibitive.
The next subsection is on propagation-based explanation methods, which are more complex to implement, but typically produce explanation results faster.

\subsection{Propagation-based explanations}
\label{secpropagation}

Algorithms using a custom back-propagation routine to create an explanation
are in stark contrast to prediction- or gradient-based explanation algorithms:
they rely on knowledge about the model's internal functioning to create more efficient or diverse explanations.

Consider gradient back-propagation that works by first decomposing a function and then performing an iterative backward mapping.
For instance, the function $f(x) = u(v(x)) = (u \circ v)(x)$ is first split into the parts $u$ and $v$ --- of which it is composed of in the first place ---
and then the gradient $\frac{\delta f}{\delta x}$ is computed iteratively $\frac{\delta f}{\delta x} = \frac{\delta u \circ v}{\delta v} \frac{\delta v}{\delta x}$ by backward mapping each component using the partial derivatives $\frac{\delta u \circ v}{\delta v}$ and $\frac{\delta v}{\delta x}$.
Similar to the computation of the gradient, all propagation-based explanations have this approach in common:
(1) each algorithm defines, explicitly or implicitly, how a network should be decomposed into different parts
and (2) how for each component the backward mapping should be performed.
When implementing an algorithm for an arbitrary network it is important to consider that different methods target different components of a network, that different decompositions for the same method can lead to different results and that certain algorithms cannot be applied to certain network structures.

For instance consider GuidedBackprop~\cite{springenberg2015striving} and Deep Taylor Decomposition~\cite[DTD]{MonPR17}.
The first targets ReLU-activations in a network and describes a backward mapping for such non-linearities, while partial derivatives are used for the remaining parts of the network.
On the other hand, DTD and many other algorithms expect the network to be decomposed into linear(izable) parts --- which can be done in several ways and may result in different explanations.

When developing such algorithms the emphasis is typically on how a backward mapping can lead to meaningful explanations, because the remaining functionality is very similar and shared across methods.
Knowing that, it is useful to split the implementation of propagation-based methods in the following two parts.
The first part contains the algorithm details--- thus defines how a network should be decomposed and how the respective mappings should be performed.
It builds upon the next part which takes care of common functionality, namely decomposing the network as previously specified and iteratively applying the mappings.
Both are denoted as "Algorithm" and "Propagation-backend" in an exemplary software stack in Figure~\ref{figsoftwarestack} in the appendix.

This abstraction has the big advantage that the complex and algorithm independent graph-processing code is shared among explanation routines
and allows the developer to focus on the implementation of the explanation algorithm itself.

We will describe in Appendix~\ref{app:backend} how a propagation backend can be implement.
Eventually it should allow the developer to realize a method in the following schematic way --- using the interface to be presented in Section~\ref{secinterface}:

\begin{pythonlisting}
# A backward mapping function, e.g., for convolutional layers
def backward_mapping(Xs, Ys, bp_Ys, bp_state):
  return compute_backward_mapping_magic()

# A class bundling all algorithm functionality
class ExplanationAlgorithm(Analyzer):
  ...
  # Defining how to perform the algorithm
  def _create_analysis(self):
    # Tell the backend that this mapping
    # should be applied, e.g., to all convolutional layers.
    register_backward_mapping(
      condition=lambda x: is_convolutional_layer(x),
      backward_mapping)
    ...

# Create and build algorithm for a model
analyzer = ExplanationAlgorithm(model)
# Perform the analysis
analyze = analyzer.analyze(x)
\end{pythonlisting}

The idea is that after decomposing the graph into layers (or sub-graphs) each layer gets assigned a mapping,
where the mappings' conditions define how they are matched.
Then the backend code will take a model and apply the explanation method accordingly to new inputs.

\subsubsection{Customizing the back-propagation}
\label{custom_backprop}

Based on the established interface we are now able to implement various propagation-based explanation methods in an efficient manner.
The algorithms will be implemented using the backend of the iNNvestigate library~\cite{alber2019innvestigate}.
Any other solution mentioned in Appendix~\ref{app:backend} could also be used.

\paragraph{Guided Backprop} 
As a first example we implement the algorithm Guided Backprop~\cite{springenberg2015striving}.
The back-propagation of Guided Backprop is the same as for the gradient computation, except that whenever a ReLU is applied in the forward pass another ReLU is applied in the backward pass.
Note that the default back-propagation mapping in iNNvestigate is the partial derivative, thus we only need to change the propagation for layers that contain a ReLU activation and apply an additional ReLU in the backward mapping.
The corresponding code looks like follows and can already be applied to arbitrary networks (see B1 in Figure~\ref{figimplemented}):

\begin{pythonlisting}
# Guidded-Backprop-Mapping
# X = input tensor of layer
# Y = ouput tensor of layer
# bp_Y = backpropagated value for Y
# bp_state = additional information on state
def guided_backprop_mapping(X, Y, bp_Y, bp_state):
  # Apply ReLU to back-propagate values
  tmp = tf.nn.relu(bp_Y)
  # Propagate back along the gradient of the forward pass
  return tf.gradients(Y, X, grad_ys=tmp)

# Extending iNNvestigate base class with the Guideded Backprop code
class GuidedBackprop(ReverseAnalyzerBase):

  # Register the mapping for layers that contain a ReLU
  def _create_analysis(self, *args, **kwargs):

    self._add_conditional_reverse_mapping(
      # Apply to all layers that contain a relu activation
      lambda layer: kchecks.contains_activation(layer, 'relu'),
      # and use the guided_backprop_mapping to do the backrop step.
      tf_to_keras_mapping(guided_backprop_mapping),
      name='guided_backprop',
    )

  return super(GuidedBackprop, self)._create_analysis(*args, **kwargs)

# Creating an instance of that analyzer
analyzer = GuidedBackprop(model_wo_sm)
# and apply it.
B1 = analyzer.analyze(x)
\end{pythonlisting}

\paragraph{Deep Taylor} Typically propagation-based methods are more evolved.
Propagations are often only described for fully connected layers and one key pattern that arises is extending this description seamlessly to convolutional and other layers.
Examples for this case are the ``Layerwise relevance propagation''~\cite{BachPLOS15}, the ``Deep Taylor Decomposition''~\cite{MonPR17} and the ``Excitation Backprop''~\cite{zhang2018top} algorithms.
Despite different motivation all algorithms yield similar propagation rules for neural networks with ReLU-activations.
The first algorithm takes the prediction values at the output neuron and calls it relevance.
Then this relevance is re-distributed at each neuron by mapping the back-propagated relevance proportionally to weights onto the inputs.
We consider here the so-called Z+ rule.
In contrast, Deep Taylor is motivated by a (linear) Taylor decomposition for each neuron and Excitation Backprop by a probabilistic ``Winner-Take-All'' scheme.
Ultimately, for layers with positive input and positive output values --- like the inner layers in VGG16 --- they all have the following propagation formula:
\begin{equation}
\begin{aligned}
  bw\_mapping(x, y, r:=bp\_y) =&\ x \odot (W^t_+ z)\\
  \text{with}\ z =&\ r \oslash (x W_+)
\end{aligned}
\end{equation}
given a fully connected layer with $W_+$ denoting the weight matrix where negative values are set to 0.
Using the library iNNvestigate this can be coded in this way:

\begin{pythonlisting}
# Deep-Taylor/LRP/EB's Z-Rule-Mapping for conv layers
# Call R=bp_Y, R for relevance
def z_rule_mapping_conv(X, Y, R, bp_state):
  # Get layer and the parameters
  layer = bp_state['layer']
  W = tf.maximum(layer.kernel, 0)

  Z = tf.keras.backend.conv2d(X, W, layer.strides, layer.padding) + b
  # normalize incoming relevance
  tmp = R / Z
  # map back
  tmp = tf.keras.backend.conv2d_transpose(
    tmp, W, (1,)+keras.backend.int_shape(X)[1:],
    layer.strides, layer.padding)
  # times input
  return tmp * X

# Extending iNNvestigate base class with the Deep Taylor/LRP/EB's Z+-rule
class DeepTaylorZ1(ReverseAnalyzerBase):
  # Register mappings for dense and convolutional layers.
  # Add Bounded DeepTaylor rule for input layer.

analyzer = DeepTaylorZ1(model_wo_sm)
B2a = analyzer.analyze(x)
\end{pythonlisting}
Unfortunately, this mapping implementation only covers 2D convolutional layers,
while other key layers like dense or other convolutional layers are not covered.
By creating another mapping for fully-connected layers (Appendix~\ref{app:deeptaylor}) the code can be applied to VGG16.
The result is shown in Figure~\ref{figimplemented} denoted as B2, where
for the constrained input layer we used the bounded rule proposed by \cite{MonPR17}.

Still, this code does not cover one-dimensional, three-dimensional or any other special type of convolutions.
Conveniently unnecessary code-replication can be avoided by using automatic differentiation.
The core idea is that many methods can be expressed as pre-/post-processing of the gradient back-propaga\-tion.
Using automatic differentiation our code example can be expressed as follows and works now with any type of convolutional layer:

\begin{pythonlisting}
# Deep-Taylor/LRP/EB's Z+-Rule-Mapping for all layers with a kernel
# Call R=bp_Y, R for relevance
def z_rule_mapping_all(X, Y, R, bp_state):
  # Get layer
  layer = bp_state['layer']
  # and create layer copy without activation part
  W = tf.maximum(layer.kernel, 0)
  layer_wo_act = kgraph.copy_layer_wo_activation(
    layer, weights=[W], keep_bias=False)

  Z = layer_wo_act(X)
  # normalize incoming relevance
  tmp = R / Z
  # map back
  tmp = tf.gradients(Z, X, grad_ys=tmp)[0]
  # times input
  return tmp * X
\end{pythonlisting}

\paragraph{LRP} For some methods it can be necessary to use different propagation rules for different layers.
E.g., Deep-Taylor requires different rules depending on the input data range~\cite{MonPR17} or
for LRP it was empirically demonstrated to be useful to apply different rules for different parts of a network.
To exemplify this, we show how to use different LRP rules for different layer types as presented in~\cite{LapAMFG17}.
In more detail, we will apply the epsilon rule for all dense layer and the alpha-beta rule for convolutional layers.
This can be implemented in iNNvestigate by changing the matching condition.
Using provided LRP-rule mappings this looks as follows:

\begin{pythonlisting}
class LRPConvNet(ReverseAnalyzerBase):

  # Register the mappings for different layer types
  def _create_analysis(self, *args, **kwargs):

    # Use Epsilon rule for dense layers
    self._add_conditional_reverse_mapping(
      lambda layer: kchecks.is_dense_layer(layer),
      LRPRules.EpsilonRule,
      name='dense',
    )
    # Use Alpha1Beta0 rule for conv layers
      self._add_conditional_reverse_mapping(
      lambda layer: kchecks.is_conv_layer(layer),
      LRPRules.Alpha1Beta0Rule,
      name='conv',
    )

  return super(LRPConvNet, self)._create_analysis(*args, **kwargs)

analyzer = LRPConvNet(model_wo_sm)
B3 = analyzer.analyze(x)
\end{pythonlisting}

The result can be examined in Figure~\ref{figimplemented} marked with B3.

\paragraph{PatternNet \& PatternAttribution} PatternNet \& PatternAttribution~\cite{kindermans2018learning} are two algorithms
that are inspired by the pattern-filter theory for linear models~\cite{haufe2014interpretation}.
They learn for each neuron in the network a signal direction called pattern.
In PatternNet the patterns are used to propagate the signal from the output neuron back to the input by iteratively using the pattern directions of the neurons
and the method can be realized with a gradient backward-pass where the filter weights are exchanged with the pattern weights.
PatternAttribution is based on the Deep Taylor Decomposition~\cite{MonPR17}.
For each neuron it searches the rootpoint in the direction of its pattern.
Given the pattern $a$ the corresponding formula is:
\begin{equation}
  bw\_mapping(x, y, r=bp\_y) = (w \odot a)^t r
\end{equation}
and it can be implemented by doing a gradient backward pass where the filter weights are element-wise multiplied with the patterns.

So far we implemented the backward-mappings as functions and registered them inside an analyzer class for backpropagation.
In the next example we will create a single class that takes a parameter, namely the patterns, and
the mapping will be a class method that uses a different pattern for each layer mapping (B4 in Figure~\ref{figimplemented}).
The following code sketches the implementations which can be found in Appendix~\ref{app:patternnet}:

\begin{pythonlisting}
# Extending iNNvestigate base class with the PatternNet algorithm
class PatternNet(ReverseAnalyzerBase):

  # Storing the patterns.
  def __init__(self, model, patterns, **kwargs):
    self._patterns = patterns[:]
    super(PatternNet, self).__init__(model, **kwargs)

  def _get_pattern_for_layer(self, layer):
    return self._patterns.pop(-1)

  # Peform the mapping
  def _patternnet_mapping(self, X, Y, bp_Y, bp_state):
    ...
    # Use patterns specific to bp_state['layer']
    ...
    
  # Register the mapping
  def _create_analysis(self, *args, **kwargs):
    ...

analyzer = PatternNet(model_wo_sm, net['patterns'])    
B4 = analyzer.analyze(x)
\end{pythonlisting}

Encapsulating the functionality in a single class allows us now to easily extend PatternNet to PatternAttribution
by changing the parameters that are used to perform the backward pass  (B5 in Figure~\ref{figimplemented}):

\begin{pythonlisting}
# Extending PatternNet to PatternAttribution
class PatternAttribution(PatternNet):

  def _get_pattern_for_layer(self, layer):
    filters = layer.get_weights()[0]
    patterns = self._patterns.pop(-1)
    return filters * patterns

analyzer = PatternAttribution(model_wo_sm, net['patterns'])
B5 = analyzer.analyze(x)
\end{pythonlisting}

\subsubsection{Generalizing to more complex networks}
\label{secgeneralization}

For our examples we relied on the VGG16 network~\cite{simonyan2014very} which is composed of linear and convolutional layers with ReLU-or Softmax-activations as well as max-pooling layers.
Recent networks in computer vision like, e.g.,  InceptionV3~\cite{szegedy2016rethinking}, ResNet50~\cite{he2016deep}, DenseNet~\cite{huang2017densenet}, or NASNet~\cite{zoph2017learning}, are far more complex and contain a variety of new layers like batch normalization layer~\cite{ioffe2015batch}, new types of convolutional layers, e.g., \cite{chollet2017xception}, and merge layers that allow for residual connections~\cite{he2016deep}.

The presented code examples either generalize to these new architectures or can be easily adapted to them.
Exemplary, Figure~\ref{figcomparenetworks} in Section~\ref{secapplications} shows a variety of algorithms applied to several state-of-the-art neural networks for computer vision.
For each algorithm the \emph{same} explanation code is used to analyze all different networks.
The exact way to adapt algorithms to new network families depends on the respective algorithm and is beyond the scope of this chapter.
Typically it consists of implementing new mappings for new layers, if required.
For more details we refer to the iNNvestigate library~\cite{alber2019innvestigate}.

\subsection{Completing the implementation}

More than the implementation of the methodological core is required to successfully apply and use explanation software.
Depending on the hyper-parameter selection and visualization approaches the explanation result may vary drastically.
Therefore it is important that software is designed to help the users to easily select the most suitable setting for their task at hand.
This can be achieved by exposing the algorithm software via an easy and intuitive interface, allowing the user to focus on the method application itself.
Subsequently we will address these topics and
as a last contribution in this subsection we will benchmark the implemented code.

\subsubsection{Interface}
\label{secinterface}

Exposing clear and easy-to-use software interfaces and routines
facilitates that a broad range of practitioners can benefit from a software package.
For instance the popular scikit-learn~\cite{pedregosa2011scikit} package offers a clear and unified interface for a wide range of machine learning methods,
which can be flexibly adjusted to more specific use cases.

In our case one commonality of all explanation algorithms is that they operate on a neural network model and therefore an interface to receive a model description is required.
There are two commonly used approaches.
The first one is chosen by several software packages, e.g., DeepLIFT~\cite{shrikumar2017learning} and the LRP-toolbox~\cite{Lapjmlr16},
and consists of expecting the model in form of a configuration (file).
A drawback of this approach is that the model needs to be serialized before the explanation can be executed.

An alternative way is to take the model represented as a memory object and operate directly with that,
e.g., DeepExplain~\cite{ancona2018towards} and iNNvestigate~\cite{alber2019innvestigate} work in this way.
Typically this memory object was build with a deep learning framework.
This approach has the advantage that an explanation can be created without additional overhead and
it is easy to use several explanation methods in the same program setup --- which is especially useful for comparisons and research purposes.
Furthermore, a model, stored in form of a configuration, can still be loaded by using the respective deep learning framework's routines and then being passed to the explanation software.

Exemplary the interface of the iNNvestigate package mimics the one of the popular software package scikit-learn and
allows to create an explanation with a few lines of code:

\begin{pythonlisting}
# Build the explanation algorithm
# with the hyper-parameter pattern_type set to 'relu'
analyzer = PatternAttribution(model_wo_sm, pattern_type='relu')
# fit the analyzer to the training data (if an analyzer requires it)
analyzer.fit(X_train)
# and apply it to an input  
e = analyzer.analyze(x)
\end{pythonlisting}

\subsubsection{Hyper-parameter selection}
\label{sechyperparameter}

\begin{figure}[t]
  \centering
  \includegraphics[width=\textwidth]{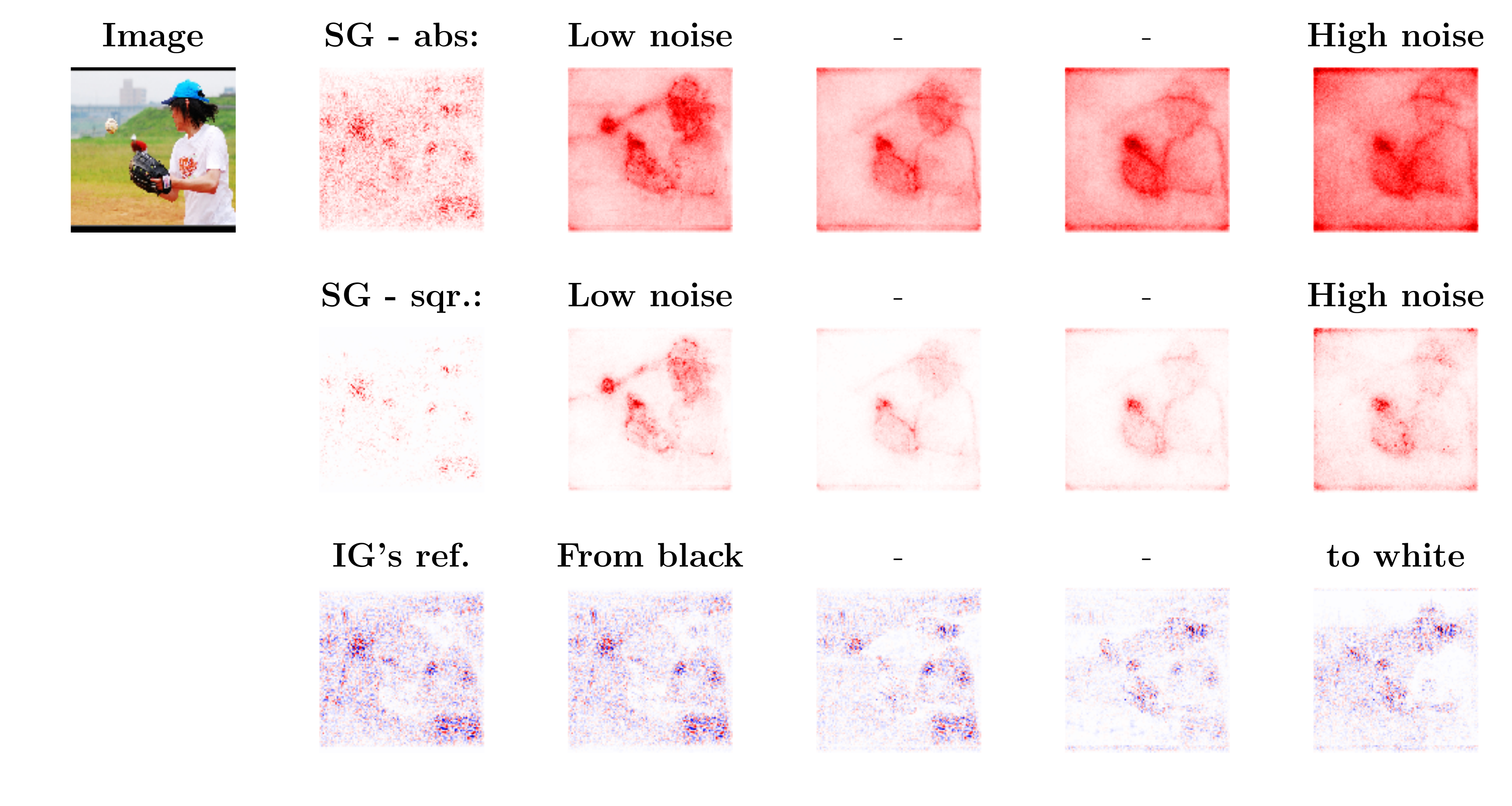}
  \caption{\textbf{Influence of hyperparamters:} Row one to three show how different hyper-parameters change the output of explanation algorithms.
    Row 1 and 2 depict the Smoothgrad (\textbf{SG}) method where the gradient is transformed into a positive value by taking the absolute or the square value respectively.
    The columns show the influence of the noise scale parameter with low to high noise from left to right.
    In row 3 we show how the explanation of the Integrated Gradients (\textbf{IG}) method varies when selecting
    as reference an image that is completely black (left side) to completely gray (middle) to completely white (right).
    Best viewed in digital and color.
  }
  \label{fighyperparameter}
\end{figure}

Like for many other tasks in machine learning explanation methods can have hyper-parameters,
but unlike for other algorithms, for explanation methods no clear selection metric exists.
Therefore selecting the right hyperparameter can be a tricky task.
One way is a (visual) inspection of the explanation result by domain experts.
This approach is suspected to be prone to the human confirmation bias.
As an alternative in image classification settings \cite{SamTNNLS17} proposed a method called ``perturbation analysis''.
The algorithm divides an image into a set of regions and sorts them in decreasing order of the ``importance'' each regions gets attributed by an explanation method.
Then the algorithm measures the decay of the neural networks prediction value when perturbing the blocks in the given order,
i.e., ``removing'' the information of the most important image parts first.
The key ideas is that if an explanation method highlights important regions better the performance will decay faster.

To visualize the sensitivity of explanation methods w.r.t. to their hyper-parameter Figure~\ref{fighyperparameter} contains two example settings.
The first example application shows the results for Integrated Gradients in row 3 where the image baseline varies from a black to a white image.
While the black, nor the white, or the gray image as reference contains any valuable information, the explanation varies significantly
--- emphasizing the need to pay attention to hyper-parameters of explanation methods.
More on the sensitivity of explanation algorithms w.r.t. to this specific parameter can be found in~\cite{kindermans2017reliability}.
Using iNNvestigate the corresponding explanation can be generated with the code in Appendix~\ref{app:hyperparameter}.

Another example is the postprocessing of the saliency output.
For instance for SmoothGrad the sign of the output is not considered to be informative
and can be transformed to a positive value by using the absolute or the square value.
This in turn has a significant impact on the result as depicted in Figure~\ref{fighyperparameter} (row 1 vs. row 2).
Furthermore, the second parameter of SmoothGrad is the scale of the noise used for smoothing the gradient.
This hyper-parameter varies from small on the left hand side to large on the right hand side
and, again, has a substantial impact on the result.
Which setting to prefer depends on the application.
The explanations were created with the code fragment in Appendix~\ref{app:hyperparameter}.

\subsubsection{Visualization}
\label{secvisualizing}

\begin{figure}[t]
  \centering
  \includegraphics[width=\textwidth]{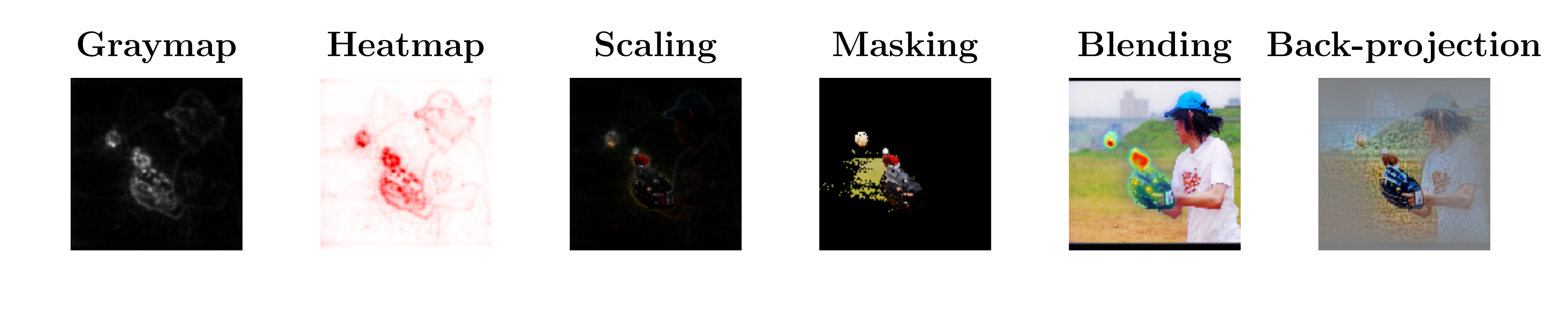}
  \caption{
    \textbf{Different visualizations:} Each column depicts a different visualization technique for the explanation of PatternAttribution or PatternNet (last column).
    The different visualization techniques are described in the text.
    Best viewed in digital and color.
  }
  \label{figvisualizing}
\end{figure}

The innate aim of explanation algorithms is to facilitate the understanding for humans.
To do so the output of algorithms needs to be transformed into a human understandable format.

For this purpose different visualization techniques were proposed in the domain of computer vision.
In Figure~\ref{figvisualizing} we depict different approaches and each one emphasizes or hides different properties of a method.
The five approaches are using
graymaps~\cite{smilkov2017smoothgrad} or single color maps to show only absolute values (column 1),
heatmaps~\cite{BachPLOS15} to show positive and negative values (column 2),
scaling the input by absolute values~\cite{sundararajan2017axiomatic} (column 3), masking the least important parts of the input~\cite{ribeiro2016should} (colmun 4),
blending the heatmap and the input~\cite{selvaraju2017grad} (column 5),
or projecting the values back into the input value range~\cite{kindermans2018learning} (column 6).
The last technique is used to visualize signal extraction techniques, while the other ones are used for attribution methods~\cite{kindermans2018learning}.
To convert color images to a two-dimensional tensor, the color channels are typically reduced to a single value by the sum or a norm.
Then the value gets projected into a suitable range and finally the according mapping is applied.
This is done for all except for the last method, which projects each value independently.
An implementation of the visualization techniques can be found in Appendix~\ref{app:visualization}.

For other domains than image classification different visualization schemes are imaginable.

\subsubsection{Benchmark}

To show the runtime efficiency of the presented code we benchmarked it.
We used the iNNvestigate library to implemente it and 
as a reference implementation we use the LRP-Caffe-Toolbox~\cite{Lapjmlr16}
because it was designed to implement algorithms with a similar complexity,
namely the LRP-variants which are the most complex algorithms we reviewed.

We test three algorithms and run them with the VGG16 network~\cite{simonyan2014very}.
Both frameworks need some time to compile the computational graph and to execute it on a batch of images,
accordingly we measure both, the setup time and the execution time, for analyzing $512$ images.

The LRP-Toolbox has a sequential and a parallel implementations for the CPU.
We show the time for the faster parallel implementation.
For iNNvestigate we evaluate the runtime on the CPU and on GPU.
The workstation for the benchmark is equipped with an Intel Xeon CPU E5-2690-v4 2.60GHz with 24 physical cores mapped to 56 virtual cores and 256G of memory.
Both implementation can use up to 32 cores. The GPU is a Nvidia P100 with 16G of memory.
We repeat each test $10$ times and report the average duration.

\begin{figure}[h]
    \centering
\begin{tikzpicture}
\begin{axis}[
    ybar,
    width=12cm,
    height=6cm,
    enlargelimits=0.15,
    legend style={at={(0.5,-0.15)},anchor=north,legend columns=-1},
    ylabel={time in s},
    symbolic x coords={Setup,Deconvnet,LRP-Epsilon,LRP-*},
    xtick=data,
    ymode=log,
    log basis y={10},
    nodes near coords,
    nodes near coords align={vertical},
    point meta=explicit symbolic,
    ]
    \addplot[color=blue,fill=blue!15] coordinates {
        (Setup, 4)
        (Deconvnet, 2550)
        (LRP-Epsilon, 2560)
        (LRP-*, 4680)
        };
    \addplot[color=red,fill=red!15] coordinates {
        (Setup, 7)[\small\text{\textcolor{black}{$0.5\times$}}\ \ \ \ \ \ ]
        (Deconvnet, 86)[\ \ \ \small\text{\textcolor{black}{$29\times$}}]
        (LRP-Epsilon, 95)[\ \ \ \small\text{\textcolor{black}{$27\times$}}]
        (LRP-*, 161)[\ \ \ \small\text{\textcolor{black}{$29\times$}}]
        };
    \addplot[color=orange,fill=orange!15] coordinates {
        (Setup, 11)[\small\text{\textcolor{black}{$0.3\times$}}\ \ \ \ \ \ ]
        (Deconvnet, 5)[\ \ \ \ \ \small\text{\textcolor{black}{$510\times$}}]
        (LRP-Epsilon, 6)[\ \ \ \ \ \small\text{\textcolor{black}{$426\times$}}]
        (LRP-*, 10)[\ \ \ \ \ \small\text{\textcolor{black}{$468\times$}}]
    };
    \legend{LRP-Toolbox (CPU)\ \ ,iNNvestigate (CPU)\ \ , iNNvestigate(GPU)};
\end{axis}
\end{tikzpicture}
    \caption{\textbf{Runtime comparison:} The figure shows the setup- and run-times for 512 analyzed images in logarithmic range for the LRP-Toolbox and the code implemented with the iNNvestigate library.
Each block contains the numbers for the setup or a different algorithm:
Deconvnet~\cite{zeiler2014visualizing}, LRP-Epsilon~\cite{BachPLOS15}, and the LRP configuration from~\cite{LapAMFG17}, denoted as LRP-*.
The numbers in black indicate the respective speedup with regard to the LRP-Toolbox.}
    \label{figruntime}
\end{figure}
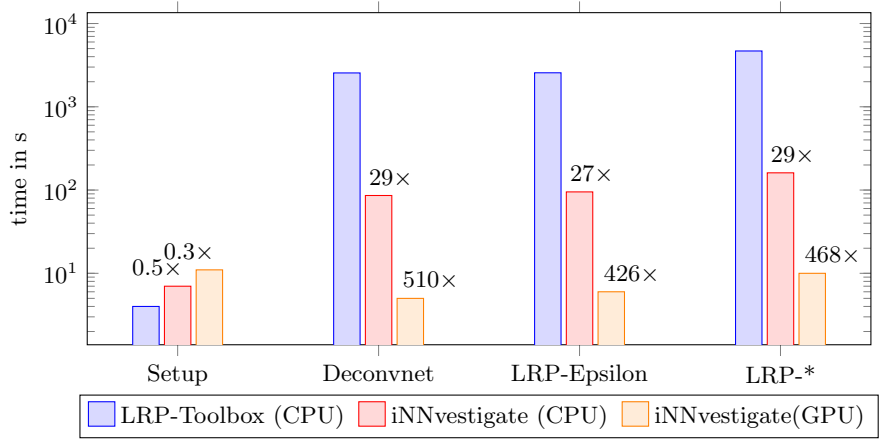

Figure~\ref{figruntime} shows the measured duration on a logarithmic scale.
The presented code implemented with the iNNvestigate library is up to $29$ times faster when both implementations run on the CPU.
This increases up to $510$ times when using iNNvestigate with the GPU compared to the LRP-Toolbox implementation on the CPU.
This is achieved while our implementations also considerably reduce the amount and the complexity of code to implement the explanation algorithms compared to the LRP-Toolbox.
On the other hand, when using frameworks like iNNvestigate one needs to compile a function graph and
accordingly the setup needs up to $3$ times as long as for the LRP-Toolbox --- yet amortizes already when analyzing a few images.

\section{Applications}
\label{secapplications}

In this section we will use the implemented algorithms and examine common application patterns for explanation methods.
For convenience we will rely on the iNNvestigate library~\cite{alber2019innvestigate} to present the following four use cases:
(1) Analyzing single (miss-)prediction to gain insights on the model, and subsequently on the data.
(2) Comparing algorithms to find a suitable explanation technique for the task at hand.
(3) Comparing prediction strategies of different network architectures.
(4) Systematically evaluating the predictions of a network.

All except for the last application, which is semi-automatic, typically require a qualitative analysis to gain insights --- and we will now see how explanation algorithms support this process.
Furthermore, this section will give a limited overview and comparison of explanation techniques.
A more detailed analysis is beyond the technical scope of this chapter.

We visualize the methods as presented in Section~\ref{secvisualizing},
i.e., use heatmaps for all methods except for PatternNet, which tries to produce a given signal and not an attribution.
Accordingly we use a projection into the input space for it.
Deconvnet and Guided Backprop are also regarded as signal extraction methods, but fail to reproduce color mappings and therefore we visualize them with heatmaps.
This allows to identify the location of signals more easily.
For more details we refer to \cite{kindermans2018learning}.

\subsection{Analyzing a prediction}

\begin{figure}[t]

  \hspace{-0.05\textwidth}
  \includegraphics[width=1.1\textwidth]{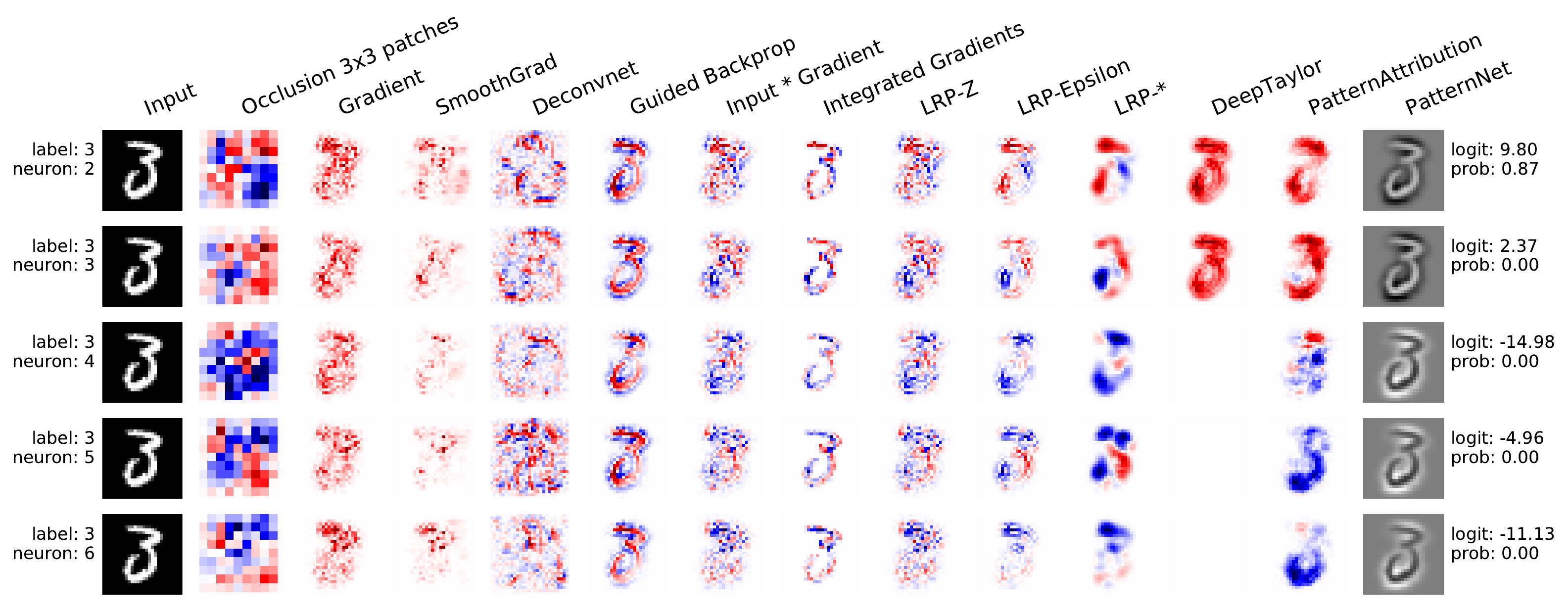}
  \caption{
    \textbf{Analyzing a prediction:} The heatmaps show different analysis for a VGG-like network on MNIST.
    The network predicts the class $2$, while the true label is $3$.
    On the left hand side the true label and for each row the respective output neuron is indicated.
    Probabilities and pre-softmax activation are denoted on the right hand side of the plot.
    Each columns is dedicated to a different explanation algorithm.
    LRP-* denotes configuration from \cite{LapAMFG17}.
    We note that Deep Taylor is not defined when the output neuron is negative.
  }
  \label{figsingleprediction}
\end{figure}

In our first example we focus on the explanation algorithms themselves and the expectations posed by the user.
Therefore we chose a dataset without irrelevant features in the input space.
In more detail we use a VGG-like network on the MNIST dataset~\cite{lecun1998mnist} with an accuracy greater than $99\%$ on the test set.

Figure~\ref{figsingleprediction} shows the result for an input image of the class $3$ that is incorrectly classified as $2$.
The different rows show the explanations for the output neurons for the classes $2$, $3$, $4$, $5$, $6$ respectively,
while each column contains the analyses of the different explanation algorithms.

The true label of the image is $3$ and also intuitively it resembles a $3$, yet it is classified as $2$.
Can we retrace why the network decided for a $2$?
Having a closer look, on the first row --- which explains the class $2$ --- the explanation algorithms suggest that the network considers the top and the left stroke as
very indicative for a $2$, and does not recognize the discontinuity between the center and the right part as contradicting.
On the other hand, a look on the second row --- which explains a $3$ --- suggests that according to the explanations the left stroke speaks against the digit being a $3$.
Potential takeaways from this are that the network does not recognize or does not give enough weight on the continuity of lines
or that the dataset does not contain enough digit $3$ with such a lower left stroke.

Taking this as an example of how such tools can help to understand a neural network,
we would like to note that all the stated points are presumptions --- based on the assumption that the explanations are meaningful.
But given this leap of faith, our argumentation seems plausible and what a user would expect an explanation algorithm to deliver.

We would also like to note that there are common indicators across different methods, e.g.,
that the topmost stroke is very indicative for a $2$
or that the leftmost stroke is not for a $3$.
This suggest that the methods base their analysis on similar signals in the network.
Yet it is not clear which method performs ``best'' and this leads us to the next example.

\subsection{Comparing explanation algorithms}

\begin{figure}[t]
 
  \hspace{-0.14\textwidth}
  \includegraphics[width=1.2\textwidth]{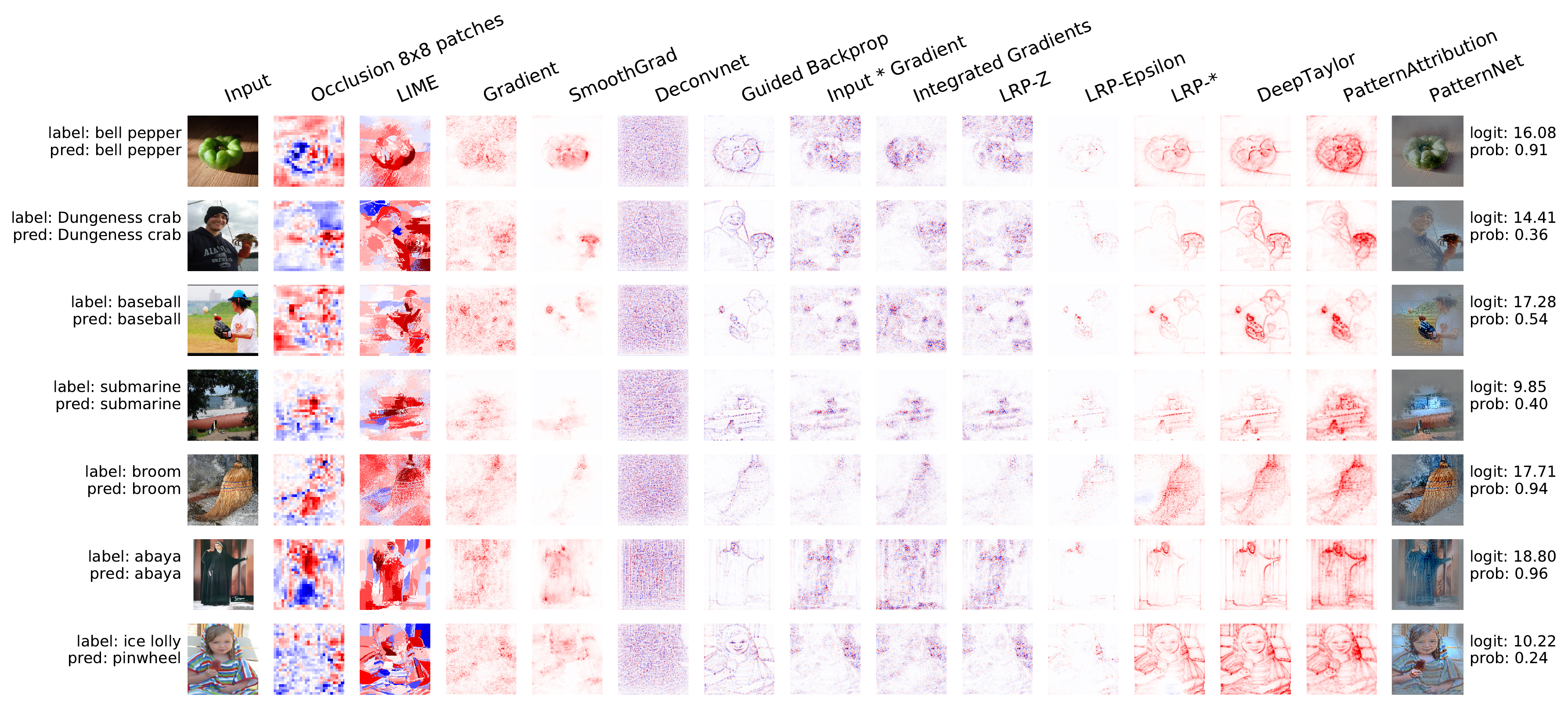}
  \caption{
    \textbf{Comparing algorithms:}
    The figure depicts the prediction analysis of a variety of algorithms (columns) for a number of input images (rows) for the VGG16 network~\cite{simonyan2014very}.
    The true and the predicted label are denoted on the left hand side and the softmax and pre-softmax outputs of the network are printed on the right hand side.
    LRP-* denotes the configuration from \cite{LapAMFG17}.
    Best viewed in digital and color.
  }
  \label{figmethodcomparison}
\end{figure}

For explanation methods there exists no clear evaluation criteria and this makes it inherently hard to find a method that ``works''
or to choose hyper-parameters (see Section~\ref{sechyperparameter}).
Therefore we argue for the need of extensive comparisons to identify a suitable method for the task at hand.

Figure~\ref{figmethodcomparison} gives an example of such a qualitative comparison and shows the explanation results for a variety of methods (columns) for a set of pictures.
We observe that compared to the previous example the analysis results are not as intuitive anymore and we also observe major qualitative differences between the methods.
For instance, the algorithms Occlusion and LIME produce distinct heatmaps compared to the other gradient- and propagation-based results.
Among this latter group, the results vary in sparseness, but also in which regions the attribution is located.
Note despite its results we added the method DeconvNet~\cite{zeiler2014visualizing} for completeness.

Consider the image in the last row, which is miss-classified as pinwheel.
While one can interpret that some methods indicate the right part of the hood as significant for this decisions, this is merely a speculation and it is hard to make sense of the analyses --- revealing the current dilemma of explanation methods and the need for more research.
Nevertheless it is important to be clear about such problems and give the user tools to make up her own opinion.

\subsection{Comparing network architectures}

\begin{figure}[t]

  \hspace{-0.145\textwidth}
  \includegraphics[width=1.2\textwidth]{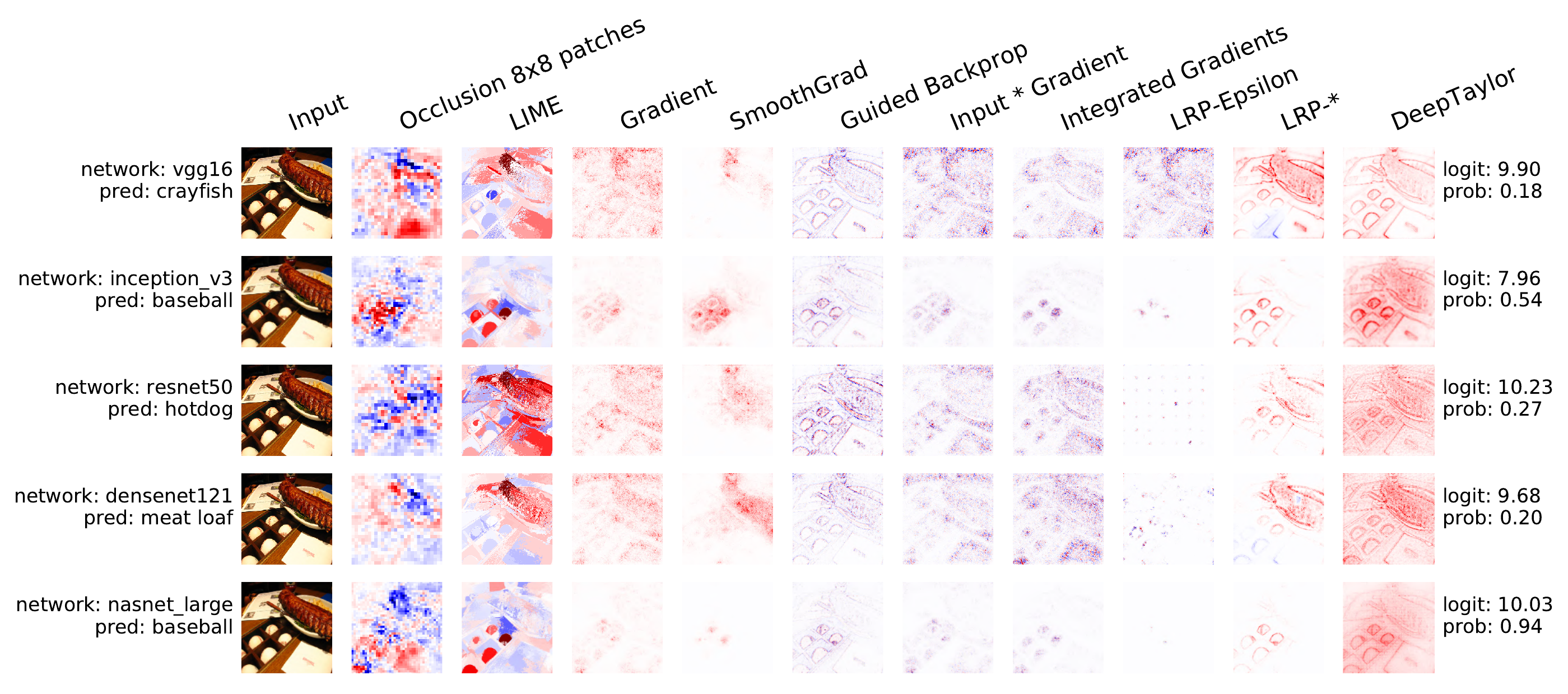}

  \caption{
    \textbf{Comparing architectures:} 
    The figure depicts the prediction analysis of a variety of algorithms (columns) for a number of neural networks (rows).
    The true label for this input image is ``baseball'' and the prediction of the respective network is given on the left hand side.
    The softmax and pre-softmax outputs of the network are printed on the right hand side.
    LRP-* denotes the configuration from \cite{LapAMFG17}.
    Best viewed in digital and color.
  }
  \label{figcomparenetworks}
\end{figure}

Another possible comparative analysis is to examine the explanations for different architectures.
This allows on one hand to assess the transferability of explanation methods and on the other hand to inspect the functioning of different networks.

Figure~\ref{figcomparenetworks} exemplarily depicts such a comparison for an image for the class ``baseball''.
We observe that the quality of the results for the same algorithm can vary significantly between different architectures,
e.g., for some algorithms the results are very sparse for deeper architectures.
Moreover, the difference between different algorithms applied to the same network seems to increase with the complexity of the architecture
(The complexity increases from the first to the last row).

Nevertheless, we note that explanation algorithms give an indication for the different prediction results of the networks
and can be a valuable tool for understanding such networks.
A similar approach can be used to monitor the learning of a network during the training.

\subsection{Systematic network evaluation}

Our last example uses a promising strategy to leverage explanation methods for analysis of networks beyond a single prediction.
We evaluate explanations for a whole dataset to search for classes where the neural network uses (correlated) background features to identify an object.
Other examples for such systematic evaluations are, e.g., grouping predictions based on their frequencies~\cite{sebastianNatComm}.
These approaches are distinctive in that they do not rely on the miss-classification as signal, i.e., one can detect undesired behavior for samples which are correctly classified by a network.

\begin{table}[h!]
  \centering
  \begin{tabular}{cc}
  \includegraphics[width=0.23\textwidth]{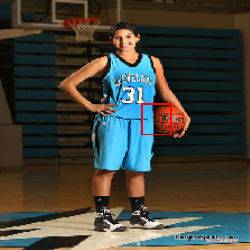}
  \includegraphics[width=0.23\textwidth]{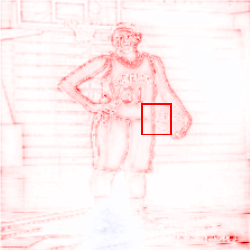} & 
  \includegraphics[width=0.23\textwidth]{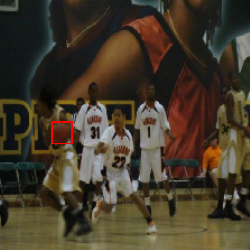}
  \includegraphics[width=0.23\textwidth]{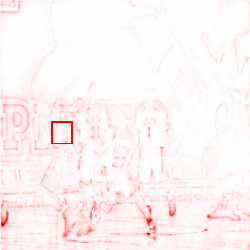} \\
  basketball & basketball \\
  \includegraphics[width=0.23\textwidth]{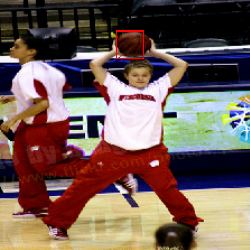}
  \includegraphics[width=0.23\textwidth]{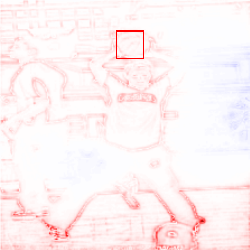} &
  \includegraphics[width=0.23\textwidth]{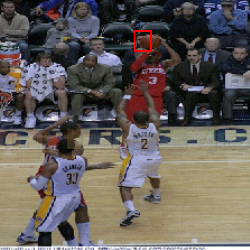} 
  \includegraphics[width=0.23\textwidth]{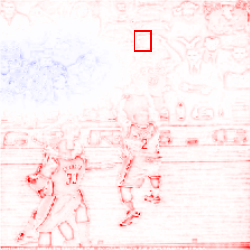} \\
  basketball & basketball \\

  \includegraphics[width=0.23\textwidth]{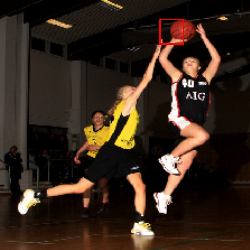}
  \includegraphics[width=0.23\textwidth]{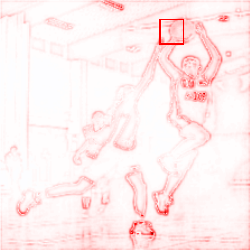} &
  \includegraphics[width=0.23\textwidth]{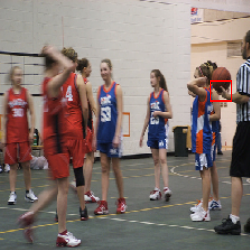}
  \includegraphics[width=0.23\textwidth]{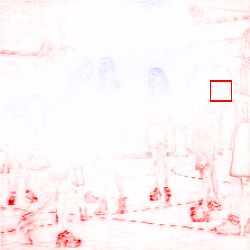} \\
  volleyball & volleyball \\

  \includegraphics[width=0.23\textwidth]{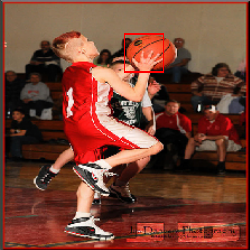}
  \includegraphics[width=0.23\textwidth]{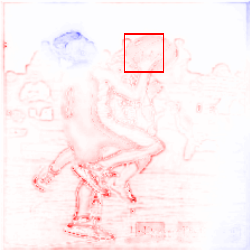} &
  \includegraphics[width=0.23\textwidth]{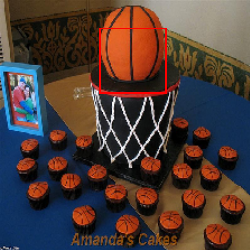}
  \includegraphics[width=0.23\textwidth]{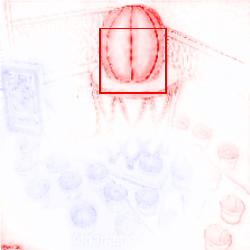} \\
  shoe shop & ping-pong ball
  \end{tabular}

  \caption{
    \textbf{Bounding box analysis:}
    The result of our bounding box analysis suggest that the target network does not use features inside the bounding box to predict the class ``basketball''.
    The images have all the true label ``basketball'' and the label beneath an image indicates the predicted class.
    We note that for none of the images the network relies on the features of a basketball for the prediction, except for the prediction ``ping-pong ball''.
    The result suggest that concept ``basketball'' is a scenery rather than a ball object for the network.
    Best viewed in digital and color.
  }
  \label{tabbboxes}
\end{table}

We use again a VGG16 network and create for each example of the ImageNet 2012~\cite{deng2009imagenet} validation set a heatmap using the LRP method with the configuration from~\cite{LapAMFG17}.
Then we compute the ratio of the attributions absolute values summed inside and outside of the bounding box, and pick the class with the lowest ration, namely ``basketball''.
A selection of images and their heatmaps is given in Table~\ref{tabbboxes}.
The first four images are correctly classified, but one can observe from the heatmaps that the network does not focus on the actual basketball inside the bounding boxes.
This suggests the suspicion that the network is not aware of the concept ``basketball'' as a ball, but rather as a scene.
Similarly, in the next three images the basket ball is not identified --- leading to wrong predictions.
Finally, the last image contains a basketball without any sport scenery and gets miss-classified as ping-pong ball.

One can argue that a sport scene is a strong indicator for the class ``basketball'',
on the other the bounding boxes make clear that the class addresses a ball rather than a scene and
the miss-classified images show that taking the scenery rather than a ball as indicator can be miss-leading.
The use of explanation methods can support developers to identify such flaws of the learning setup caused by, e.g., biased data or networks that rely on the ``wrong'' features~\cite{sebastianNatComm}.

\section{Software packages}
\label{secsoftwarepackages}

In this section we would like to give an overview on software packages for explanation techniques.

Accompanying the publication of algorithms many authors released also dedicated software.
For the LRP-algorithm a toolbox was published~\cite{Lapjmlr16} that contains explanatory code in Python and MatLab as well as a faster Caffee-implementation for production purposes.
For the algorithms DeepLIFT~\cite{shrikumar2017learning}, DeepSHAPE~\cite{lundberg2017unified}, "prediction difference  analysis"~\cite{zintgraf2017visualizing}, and LIME~\cite{ribeiro2016should} the authors also published source code that is based on Keras/Tensorflow, Tensorflow, Tensorflow and scikit-learn respectively.
For the algorithm GradCam~\cite{selvaraju2017grad} the authors published a Caffee-based implementation.
There exist more GradCam implementations for other frameworks, e.g.,~\cite{raghakotkerasvis}.

Software packages that contain more than one algorithm family are the following.
The software to the paper DeepExplain~\cite{ancona2018towards} contains implementations for the gradient-based algorithms saliency map, gradient * input, Integrated Gradients, one variant of DeepLIFT and LRP-Epsilon as well as for the occlusion algorithm.
The implementation is based on Tensorflow.
The Keras-based software keras-vis~\cite{raghakotkerasvis} offers code to perform activation maximization, saliency algorithms Deconvnet and GuidedBackprop as well as GradCam.
Finally, the library iNNvestigate~\cite{alber2019innvestigate} is also Keras-based and contains implementations for the algorithms saliency map, gradient * input, Integrated Gradients, Smoothgrad, DeconvNet, GuidedBackprop, Deep Taylor Decomposition, different LRP algorithms as well as PatternNet and PatternAttribution.
It also offers an interface to facilitate the implementation of propagation-based explanation methods.

\section{Challenges}
\label{secchallenges}

Neural networks come in a large variety.
They can be composed of many different layers and be of complex structure (e.g., Figure~\ref{fignasnetblocks} shows the sub-blocks of the NASNetA network).
Many (propagation-based) explanation methods are designed to handle fully connected layers in the first place, 
yet to be universally applicable a method and its implementations must be able to scale beyond fully-connected networks and be able to generalize to new layer types.
In contrast, the advantage of methods that only use a model's prediction or gradient is their applicability independent of a network's complexity,
yet they are typically slower and cannot take advantage of high level features like propagation methods~\cite{LapAMFG17,selvaraju2017grad}.

To promote research on propagation methods for complex neural networks it is necessary alleviate researchers from unnecessary implementation efforts.
Therefore it is important that tools exist that allow for fast prototyping and let researchers focus on algorithmic developments.
One example is the library iNNvestigate, which offers an API that allows to modify the backpropagation easily and
implementations of many of state-of-the explanation methods ready for advanced neural networks.
We showed in Section~\ref{secgeneralization} how a library like iNNvestigate helps to generalize algorithms to various architectures.
Such efforts are promising to facilitate research as they make it easier to compare and develop methods as well as
facilitate faster adaption to (recent) developments in deep learning.

For instance, despite first attempts~\cite{ArrWASSA17,poener18} LSTMs~\cite{hochreiter1997long,sutskever2014sequence} and attention layers~\cite{vaswani2017attention} are still a challenge for most propagation-based explanation methods.
Another challenge are architectures discovered automatically with, e.g., neural architecture search~\cite{zoph2017learning}.
They often outperform competitors that were created by human intuition, but are very complex.
A successful application of and examination with explanation methods can be a promising way to shed led into their workings.
The same reasoning applies to networks like SchNet~\cite{schutt2017schnet}, WaveNet~\cite{van2016wavenet}, and AlphaGo~\cite{silver2016mastering} --- which led to breakthroughs in their respective domains
and a better understanding of their predictions would reveal valuable knowledge.

Another open research question regarding propagation-based methods concerns the decomposition of network into components.
Methods like Deep Taylor Decomposition, LRP, DeepLIFT, DeepSHAPE decompose the network and create an explanation based on the linearization of the respective components.
Yet networks can be decomposed in different ways:
for instance the sequence of a convolutional and a batch normalization layer can be treated as two components or be represented as one layer where both are fused.
Another example is the treatment of a single batch normalization layer which can be seen as one or as two linear layers.
Further examples can be found and it is not clear how the different approaches to decompose a network influence the result of the explanation algorithms and requires research.

\section{Conclusion}
\label{secconclusion}

Explanation methods are a promising approach to leverage hidden knowledge about the workings of neural networks,
yet the complexity of many methods can prevent practitioners from implementing and using them for research or application purposes.
To alleviate this shortcoming it is important that accessible and efficient software exists.
With this in mind we explained how such algorithms can be implemented efficiently by using deep learning frameworks like TensorFlow and Keras
and showcased important algorithm and application patterns.
Moreover, we demonstrated different exemplary use cases of explanation methods such as examining miss-classifications, comparing algorithms, and detecting if a network focuses on the background.
By building such software the field will hopefully be more accessible for non-experts and find appeal in the broader sciences.
We also hope that it will help researchers to tackle recent developments in deep learning.

\section*{Acknowledgment}
The authors thank Sebastian Lapuschkin, Gr{\'e}goire Montavon and Klaus-Robert M\"uller for their valuable feedback.
This work was supported by the Federal Ministry of Education and Research (BMBF) for the Berlin Big Data Center 2 - BBDC 2
(01IS18025A).


%
%
%
\bibliographystyle{splncs04.bst}
\bibliography{main.bib}

\newpage
\appendix

\section{Section 2 - supplementary content}

\subsection{Propagation backend}
\label{app:backend}

\begin{figure}[t]
  \centering
  \includegraphics[width=0.5\textwidth]{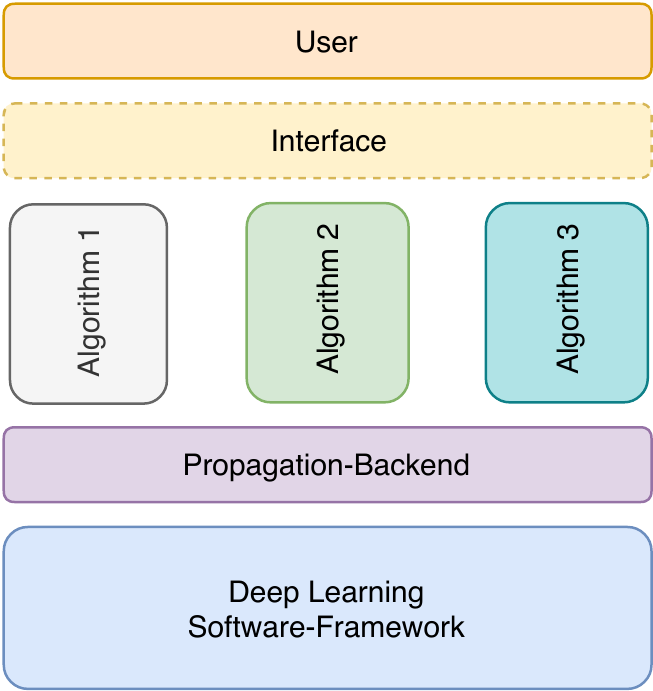}
  \caption{\textbf{Exemplary software-stack:} The diagram depicts exemplary the software stack of iNNvestigate~\cite{alber2019innvestigate}.
  It shows how different propagation-based methods are build on top of a common graph-backend and expose their functionality through a common interface to the user.
  }
  \label{figsoftwarestack}
\end{figure}

\subsubsection{Creating a propagation backend}
\label{secbackpropbackend}
Let us reiterate the aim, which is to create routines that capture common functionality to all propagation-based algorithms and thereby facilitate their efficient implementation.
Given the information which graph-parts shall be mapped and how,
the backend should decompose the network accordingly and then process the back-propagation as specified.
It would be further desirable that the backend is able to identify if a given neural network is not compatible with an algorithm, e.g., because the algorithm does not cover certain network properties.

In this regard we see as major challenges for creating an efficient backend the following:
\begin{description}
\item[Interface:] How shall an algorithm specify the way a network should be decomposed and how should each backward mapping be performed?
\item[Graph matching:] Decomposing the neural network according to the algorithm's specifications and, ideally, detecting possible incompatibilities. Note that the specifications can describe the structure of the targeted components as well as their location in the network,
e.g., DTD treats layer differently depending where they are located in the network.
\item[Back-propagation:] Once determined which backward mapping is used for which part of the network graph, the respective mappings should be applied in the right order until the final explanation is produced.
\end{description}

The first two challenges are solved by choosing appropriate abstractions.
The abstractions should be fine-grained enough to enable the implementation of a wide range of algorithms, while being coarse-grained enough to allow for an efficient implementation.
The last challenge is in the first place an engineering task.

\paragraph{Interface \& Matching}
\label{secinterfaceandmatching}

\begin{figure}[t]
  \centering
  \includegraphics[width=\textwidth]{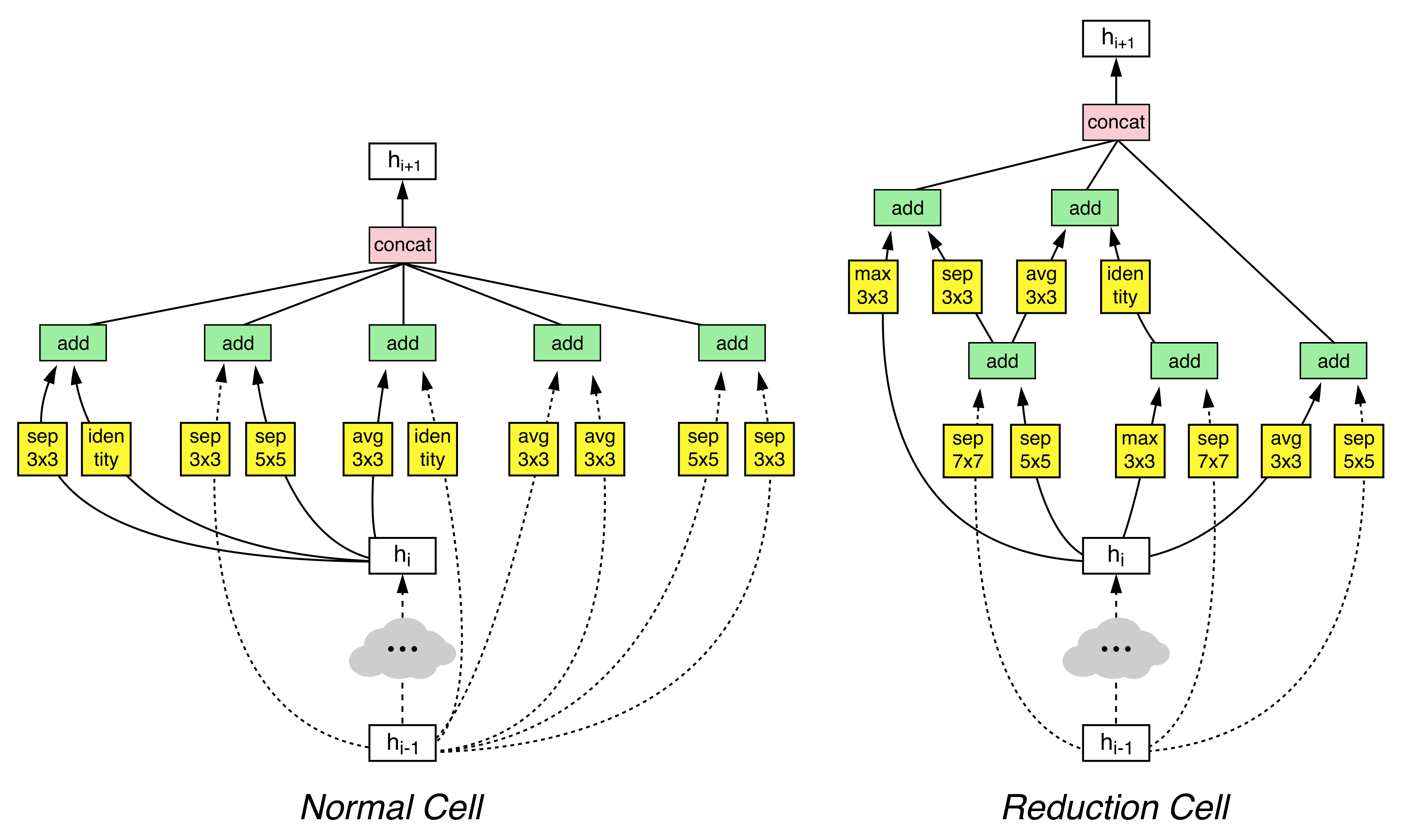}
  \caption{\textbf{NASNetA cells:} The computer vision network NASNetA~\cite{zoph2017learning} was created with automatic machine learning, i.e., the architecture of the two depicted building blocks was found with an automated algorithm.
  The normal cell and the reduction cell have the same purpose as convolutional or max-pooling layers in other networks, but are far more complex. (Figure is from~\cite{zoph2017learning}.)
  }
  \label{fignasnetblocks}
\end{figure}

The first step towards a clear interface is to regard a neural network as a directed-acyclic-graph (DAG) of layers --- instead of a stack of layers.
The notion of a graph of "layers" might not seem intuitive in the first place and comes from the time when neural networks were typically sequential, thus one layer was stacked onto another.
Modern networks, e.g., as NASNetA in Figure~\ref{fignasnetblocks}, can be more complex and in such architectures each layer is rather a node in a graph than a layer in a stack.
Regardless of that, nodes in such a DAG are still commonly called layers and we will keep this notation.

A second step is to be aware of DAG's granularity.
Different deep learning frameworks represent neural networks in different ways and layers can be composed of more basic operations like additions and dot products, which in turn can be decomposed further.
The most intuitive and useful level for implementing explanation methods is to view each layer as node.
A more fine-grained view is in many cases not needed and would only complicate the implementation.
On the other hand, we note that it might be desired or necessary to fuse layers of networks into one node, e.g., adjacent convolutional and batch normalization layers can be expressed as a single convolutional layer.

Building on this network representation, there are two interfaces to define. One to define where a mapping shall be applied and one how it should be performed. 

There are two ways to realize the matching interface and they can be sketched as follows.
The first binds a custom backward mapping before or during network building to a method of a layer class --- statically by extending a layer class or by overloading its gradient operator.
The second receives the already build model and matches the mappings dynamically to the respective layer nodes.
This can be done by evaluating a programmable condition for each layer instance or node in order to assign a mapping.
Except for the matching conditions, both techniques expose the same interface and in contrast to the first approach the later is more challenging to implement, but has several advantages:
(1) It exposes a clear interface by separation of concerns: the model building happens independently of the explanation algorithm.
(2) The forward DAG can be modified before the explanation is applied, e.g., batch normalization layer can be fused with convolutional layers.
(3) When several explanation algorithms are build for one network, they can share the forward pass.
(4) The matching conditions can introspect the whole model, because the model was already build at that point in time.
(5) One can build efficiently the gradient w.r.t. explanations by using forward-gradient computation --- in the background and for all explanation algorithms by using automatic differentiation.

The two approaches can be sketched in Python as follows:

\begin{pythonlisting}
# Approach A 
# Use mapping Y for layer type X
register_mapping_for_layer_type(layer_type_X, mapping_Y)
build_model_with_custom_mapping()
execute_explanation()

# Approach B
model = build_model()
graph = extract_and_update_graph(model)
for node in graph:
  # Match node to mapping based on conditions
  # A node can be a layer or a sub-graph.
  # Condition can introspect whole model for decision.
  mapping = match_node_to_mapping(node, model.graph)
  assign_mapping_to_node(node, mapping)
\end{pythonlisting}

The second interface addresses the backward mapping and is a function that takes as parameters the input and output tensors of the targeted layer, the respective back-propagated values for the output tensors and, optionally, some meta-information on the back-propagation process.
The following code segment shows the interface of a backward mapping function in the iNNvestigate library.
Due to same purpose other implementations have very similar interfaces.
\begin{pythonlisting}
# Xs = input tensors of a layer or sub-graph
# Ys = ouput tensors of a layer or sub-graph
# bp_Ys = back-propagated values for Ys
# bp_state = additional information on state
# return back-propagated values for Xs
def backward_mapping(Xs, Ys, bp_Ys, bp_state):
  # the backward mapped tensors correspond in shape
  # with respective the output tensors of the forward pass
  assert len(Ys) == len(bp_Ys)
  assert all(Y.shape == bp_Y.shape for Y, bp_Y in zip(Ys, bp_Ys))

  bp_Xs = compute_backward_mapping_magic()

  # the returned tensors correspond in shape
  # with the respective input tensors of the forward pass
  assert len(Ys) == len(bp_Ys)
  assert all(Y.shape == bp_Y.shape for Y, bp_Y in zip(Ys, bp_Ys))
  return bp_Xs
\end{pythonlisting}
Note that this signature can not only be used for the backward mapping of layers, but for any connected sub-graph.
In the remainder we will use a simplified interface where each layer has only one input and one output tensor.

\paragraph{Back-propagation}
Having matched backward mappings with network parts the backend still needs to create the actual backward propagation.
Practically this can be done explicitly, as we will show below, or by overloading the gradient operator in the deep learning framework of choice.
While the latter is easier to implement it less flexible and has the dis-advantages mentioned above.

The implementations of neural networks is characterized by their layer-oriented structure and
the simplest of them are sequential neural networks where each layer is stacked on another layer.
To back-propagate through such a network one starts with the model's output value and
propagates from top layer to the next lower one and so on.
Given mapping functions that take a tensor and back-propagate along a layer, this can be sketched as follows:

\begin{pythonlisting}
current = output
for layer in model.layers[::-1]:
  current = back_propagate(layer.input, layer.output, current)
analysis = current
\end{pythonlisting}

In general neural networks can be much more complex and are represented as directed, acyclic graphs.
This allows for multiple input and output tensors for each "layer node".
An efficient implementation is for instance the following.
First the right propagation order is established using the depth-first search algorithm to create a topological ordering~\cite{cormen2009introduction}.
Then given this ordering, the propagation starts at the output tensors and proceeds in direction of the input tensors.
At each step, the implementation collects the required inputs for each node, applies the respective mapping and keeps track of the back-propagated tensors after the mapping.
Note, nodes that branch in the forward pass, i.e., have an output tensor that is used several times in the forward pass, receive several tensors as inputs in the backward pass.
These need to be reduced to a single tensor before being fed to the backward mapping.
This is typically like in the gradient computation, namely by summing the tensors:

\begin{pythonlisting}
intermediate_tensors = {output: output}
execution_order = calculate_execution_order()
for layer, inputs, outputs in execution_order[::-1]:
  # gather corresponding back-propagated tensors for each output tensor
  back_propagated_values = [
    # Reduce to single tensor if the forward passed branched!
    sum(intermediate_tensors[t])
    for t in outputs
  ]

  # backprop through layer
  tmp = back_propagate(inputs, outputs, back_propagated_values)

  # store intermediate tensors
  for input, intermediate in zip(inputs, tmp):
    if input in intermediate_tensors:
      intermediate_tensors[input] = [intermediate]
    else:
      # The corresponding forward tensor branched!
      intermediate_tensors[input].append(intermediate)

# get the last output
analysis = intermediate_tensors[model.input]  
\end{pythonlisting}

Despite its relative simplicity, implementing and debugging such an algorithm can be tedious.
This among propagation-based methods common operation is part of the iNNvestigate library and
as a result one only needs to specify how the back-propagation through specific layers should be performed.
Even handier, as default backward mapping the gradient-propagation is used and one only needs to specify whenever the back-propagation should be performed differently.

\subsection{Deep Taylor}
\label{app:deeptaylor}

The Deep Taylor mapping for dense layers:

\begin{pythonlisting}
# Deep-Taylor/LRP/EB's Z+-Rule-Mapping for dense layers
# Call R=bp_Y, R for relevance
def z_rule_mapping_dense(X, Y, R, bp_state):
  # Get layer and the parameters
  layer = bp_state['layer']
  W = tf.maximum(layer.kernel, 0)

  Z = tf.tensordot(X, W, 1) + b
  # normalize incoming relevance
  tmp = R / Z
  # map back
  tmp = tf.tensordot(tmp, tf.transpose(W), 1)
  # times input
  return tmp * X
\end{pythonlisting}

\subsection{PatternNet}
\label{app:patternnet}

The exemplary implementation for PatterNet discussed in Section~\ref{secpropagation}:

\begin{pythonlisting}
# Extending iNNvestigate base class with the PatternNet algorithm
class PatternNet(ReverseAnalyzerBase):

  # Storing the patterns.
  def __init__(self, model, patterns, **kwargs):
    self._patterns = patterns[:]
    super(PatternNet, self).__init__(model, **kwargs)

  def _get_pattern_for_layer(self, layer):
    return self._patterns.pop(-1)

  def _patternnet_mapping(self, X, Y, bp_Y, bp_state):
    # Get layer,
    layer = bp_state['layer']
    # exchange kernel weights with patterns,
    weights = layer.get_weights()
    weights[0] = self._get_pattern_for_layer(layer)
    # and create layer copy without activation part and patterns as filters
    layer_wo_act = kgraph.copy_layer_wo_activation(layer, weights=weights)

    if kchecks.contains_activation(layer, 'relu'):
      # Gradient of activation layer
      tmp = tf.where(Y > 0, bp_Y, tf.zeros_like(bp_Y))
    else:
      # Gradient of linear layer
      tmp = bp_Y

    # map back along layer with patterns instead of weights
    pattern_Y = layer_wo_act(X)
    return tf.gradients(pattern_Y, X, grad_ys=tmp)[0]
    
  # Register the mappings
  def _create_analysis(self, *args, **kwargs):
    self._add_conditional_reverse_mapping(
      # Apply to all layers that contain a kernel
      lambda layer: kchecks.contains_kernel(layer),
      tf_to_keras_mapping(self._patternnet_mapping),
      name='pattern_mapping',
  )
  return super(PatternNet, self)._create_analysis(*args, **kwargs)

analyzer = PatternNet(model_wo_sm, net['patterns'])    
B4 = analyzer.analyze(x)
\end{pythonlisting}

\subsection{Hyper-paramter selection}
\label{app:hyperparameter}

The exemplary hyper-parameter selection for Integrated Gradients:
\begin{pythonlisting}
IG = []
# Take 5 samples from network's input value range
for ri in np.linspace(net['input_range'][0], net['input_range'][1], num=5):
  # and analyze with each.
  analyzer = innvestigate.create_analyzer(
    'integrated_gradients',
    model_wo_sm,
    reference_inputs=ri,
    steps=32
  )
  IG.append(analyzer.analyze(x))
\end{pythonlisting}

The exemplary hyper-parameter selection for SmoothGrad:

\begin{pythonlisting}
SG1, SG2 = [], []
# Take 5 scale samples for the noise scale of smoothgrad.
for scale in range(5):
  noise_scale = (net['input_range'][1]-net['input_range'][0]) * scale / 5
  # Smoothgrad with absolute gradients
  analyzer = innvestigate.create_analyzer(
    'smoothgrad',
    model_wo_sm,
    augment_by_n=32,
    noise_scale=noise_scale,
    postprocess='abs'
  )
  SG1.append(analyzer.analyze(x))

  # Smoothgrad with with squared gradients
  analyzer = innvestigate.create_analyzer(
    'smoothgrad',
    model_wo_sm,
    augment_by_n=32,
    noise_scale=noise_scale,
    postprocess='square'
  )
  SG2.append(analyzer.analyze(x))
\end{pythonlisting}

\subsection{Visualization}
\label{app:visualization}

The exemplary implementation of visualization approaches discussed in Section~\ref{secvisualizing}:

\begin{pythonlisting}
def explanation_to_heatmap(e):
  # Reduce color axis
  tmp = np.sum(e, axis=color_channel_axis)
  # To range [0, 255]
  tmp = (tmp / np.max(np.abs(tmp))) * 127.5 + 127.5
 
  # Create and apply red-blue heatmap
  colormap = matplotlib.cm.get_cmap("seismic")
  tmp = colormap(tmp.flatten().astype(np.int64))[:, :3]
  tmp = tmp.reshape(e.shape)
  return tmp

def explanation_to_graymap(e):
  # Reduce color axis
  tmp = np.sum(np.abs(e), axis=color_channel_axis)
  # To range [0, 255]
  tmp = (tmp / np.max(np.abs(tmp))) * 255
 
  # Create and apply red-blue heatmap
  colormap = matplotlib.cm.get_cmap("gray")
  tmp = colormap(tmp.flatten().astype(np.int64))[:, :3]
  tmp = tmp.reshape(e.shape)
  return tmp

def explanation_to_scale_input(e):
  # Create scale
  e = np.sum(np.abs(e), axis=color_channel_axis, keepdims=True)
  scale = e / np.max(e)
 
  # Apply to image
  return (x_not_preprocessed / 255) * scale

def explanation_to_mask_input(e):
  # Get highest scored segments
  # Segments are reused from the LIME example.
  segments_scored = [(np.max(e[0][segments == sid]), sid) for sid in range(nr_segments)]
  highest_ones = sorted(segments_scored, reverse=True)[:50]
  
  # Compute mask
  mask = np.zeros_like(segments)
  for _, sid in highest_ones:
    mask[segments == sid] = 1
  
  # Apply mask
  ret = (x_not_pp.copy() / 255)
  ret[0][mask == 0] = 0
  return ret

def explanation_to_blend_w_input(e):
  e = np.sum(np.abs(e), axis=channel_axis, keepdims=True)
  # Add blur
  e = skimage.filters.gaussian(x[e], 3)[None]
  # Normalize
  e = (e - e.min())/(e.max()-e.min())
  # Get and apply colormap
  heatmap = plot.get_cmap("jet")(e[:, :,:,0])[:,:,:,:3]
  # Overlap
  ret = (1.0-e) * (x_not_pp / 255) + e * heatmap
  return ret

def explanation_to_projection(e):
  # To range [0, 1]
  return (e / np.max(np.abs(e))) + 0.5
\end{pythonlisting}

\end{document}